\documentclass[10pt,twocolumn,letterpaper]{article}

\usepackage{iccv}
\usepackage{times}
\usepackage{epsfig}
\usepackage{graphicx}
\usepackage{amsmath}
\usepackage{amssymb}

\usepackage{multirow}

\usepackage[breaklinks=true,bookmarks=false]{hyperref}

\iccvfinalcopy 

\def\iccvPaperID{****} 

\ificcvfinal\fi

\begin{document}

\title{LODE: Deep Local Deblurring and A New Benchmark}

\author{Zerun Wang$^{1}$\thanks{Equal Contribution}, Liuyu Xiang$^{1*}$, Fan Yang$^{1}$, Jinzhao Qian$^{1}$, Jie Hu$^{2}$, Haidong Huang$^{2}$, \\ Jungong Han$^{3}$, Yuchen Guo$^{1}$\thanks{Corrresponding Author}, Guiguang Ding$^{1 \dag} $ \\
$^1$ BNRist, School of Software, Tsinghua University, Beijing, China 
$^2$ OPPO Inc. Guangdong, China \\
$^3$ Computer Science Department, Aberystwyth University, SY23 3FL, UK}


\maketitle
\ificcvfinal\thispagestyle{empty}\fi

\begin{abstract}

While recent deep deblurring algorithms have achieved remarkable progress, most existing methods focus on the global deblurring problem, where the image blur mostly arises from severe camera shake. We argue that the local blur, which is mostly derived from moving objects with a relatively static background, is prevalent but remains under-explored. In this paper, we first lay the data foundation for local deblurring by constructing, for the first time, a \textbf{lo}cal-\textbf{de}blur (LODE) dataset consisting of 3,700 real-world captured locally blurred images and their corresponding ground-truth. Then, we propose a novel framework, termed \textbf{bl}ur-\textbf{a}ware \textbf{de}blurring network (BladeNet), which contains three components: the Local Blur Synthesis module generates locally blurred training pairs, the Local Blur Perception module automatically captures the locally blurred region and the Blur-guided Spatial Attention module guides the deblurring network with spatial attention. This framework is flexible such that it can be combined with many existing SotA algorithms. We carry out extensive experiments on REDS and LODE datasets showing that BladeNet improves PSNR by $\mathbf{2.5}$ \textbf{dB} over SotAs for local deblurring while keeping comparable performance for global deblurring. We will publish the dataset and codes.

\end{abstract}

\vspace{-5pt}
\section{Introduction}



\begin{figure}
\begin{center}
\includegraphics[width= \linewidth]{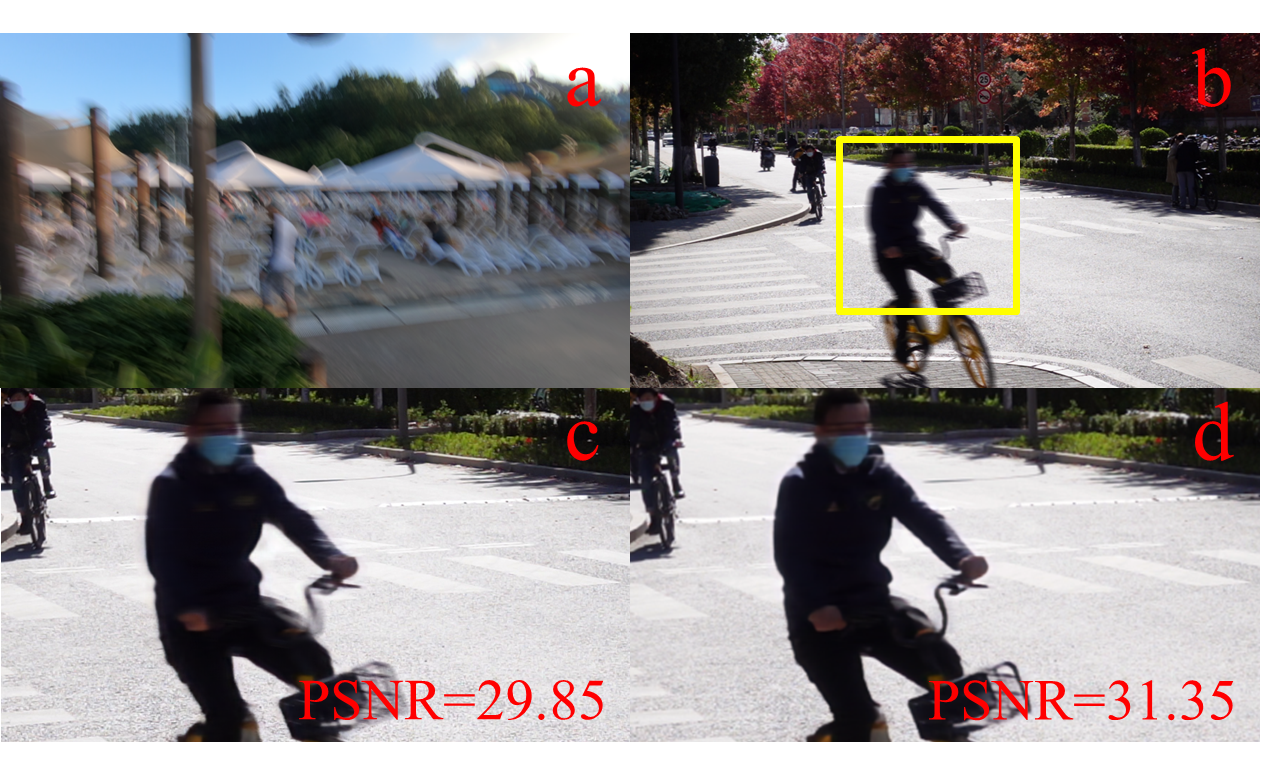}
\end{center}
\vspace{-10pt}
\caption{(a) Global blur by camera moving from REDS~ \cite{Nah_2019_CVPR_Workshops_REDS}. (b) Local blur by object moving from LODE. (c) A SotA global deblurring approach~\cite{zhang2019deep} only has $29.85dB$ PSNR on a local-blur image. (d) Our approach achieves $31.35dB$ PSNR with a much clearer result. Zoom in for better view.}
\label{fig:intro_fig}
\vspace{-5pt}
\end{figure}

How to improve the quality of images is an active research topic in recent decades, especially with the emergence of digital imaging devices~\cite{koh2020single,mafi2019comprehensive}. Among all issues, image blur is one of the commonest flaws in real-world cases, which is typically caused by moving cameras or objects.
Before the era of deep learning, many deblurring approaches had been proposed to solve the problem by estimating blur kernels explicitly~\cite{whyte2012non,gupta2010single}. The assumption behind is that there are well formulated blur kernels that turn a sharp image into a blurred counterpart. Unfortunately, the blur kernel usually fails to model the real-world blur caused by complex motion.
Recently, the emergence of deep convolutional neural networks (CNNs)~\cite{lecun2015deep} has made it possible to learn from data automatically without specifying the underlying blur kernels. With its powerful feature learning capability, deep CNN-based deblurring approaches~\cite{nah2017deep,kupyn2018deblurgan,kupyn2019deblurgan} achieves significant improvement over the non-deep ones. 

Despite the tremendous progress it has made, we argue that most previous deep learning based methods focus on \textit{global deblurring}, in which the global blur mostly caused by camera shake is dominant (Figure \ref{fig:intro_fig} (a)). The \textit{local deblurring} problem, where the local blur mostly caused by moving objects is dominant (Figure \ref{fig:intro_fig} (b)), is still under-explored. The global blur typically happens when one captures images with \textit{moving cameras} while the local blur mostly happens when one snapshots \textit{moving objects} with relatively static cameras. It is worth noting that in real-world scenarios, blurred images usually contain mixed types of blur, and we distinguish global and local deblurring problems by looking at which type of blur is dominant. We also conduct statistical analysis (Section 5.1) to show the difference between them. 

The absence of deep local deblurring investigation is twofolded: first, existing deblurring datasets such as GOPRO \cite{nah2017deep} and REDS \cite{Nah_2019_CVPR_Workshops_REDS} are captured with severe and deliberate camera shake, making most of the images globally blurred (Figure \ref{fig:intro_fig} (a)). Second, most deep models neglect the characteristics of local blur.
More importantly, we discover that these two tasks are intrinsically different such that a successful global deblurring approach may perform poorly for local deblurring. 
As illustrated in Figure \ref{fig:intro_fig} (c-d), SotA global deblurring method SDNet ~\cite{zhang2019deep} achieves only $29.85dB$ in PSNR while our proposed local deblurring approach yields $31.35dB$ in PSNR with a much sharper result. 


The local blur is ubiquitous in daily life. For example, a surveillance camera is usually static while the objects of interest (say, a moving car or person) are moving fast, or one uses a cellphone to snapshot a key moment in a sports game. The broad occurrence of local blur and the poor performance of current deblurring approaches for this task motivate us to investigate local deblurring deeper. In particular, we address this issue from data and model perspectives. Data is the foundation of deep learning based approaches. 
Since existing datasets \cite{nah2017deep,Nah_2019_CVPR_Workshops_REDS} fail to facilitate relevant studies on local deblurring, we construct a large-scale dataset for \textbf{lo}cal \textbf{de}blurring (LODE) by high-frame-rate cameras. We capture moving objects with a relatively static camera. This way ensures that the image is locally blurred since the background is relatively clear while the moving objects are blurred. LODE has $3,700$ (and is still increasing) blurred images from various scenes, together with ground truth (sharp images captured in high frame rate). To the best of our knowledge, LODE is the first dataset for deep local deblurring. 

From the model perspective, we correspondingly propose a novel \textbf{bl}ur-\textbf{a}ware \textbf{de}blurring network (BladeNet) where a novel local blur synthesis (LBS) module for local blurred training data generation is presented. It utilizes an auxiliary dataset equipped with object segmentation masks to generate synthetic images with local blur. This procedure enables training of a local deblurring model even when local blurry training images are unavailable. In addition, we further introduce a local blur perception (LBP) module and a blur-guided spatial attention (BSA) module. The blur perception module captures the blur region and predicts the probability map of the blurred area, such that local blur areas could be restored and backgrounds are kept sharp. The blur-guided spatial attention module uses the segmentation mask trained in the data synthesis module as a supervision signal to guide the spatial-attention to focus on those locally blurred areas. Both modules contribute to the performance improvement for local deblurring. Based on BladeNet, the PSNR is improved by about $\mathbf{2.5}$ \textbf{dB} for the local deblurring task, compared to the SotAs for deblurring, while still comparable to them for global deblurring. In summary, our contributions are fourfolds:

1. We demonstrate that \textit{local deblurring} is intrinsically different from \textit{global deblurring}. The successful SotAs for global deblurring may perform poorly for local deblurring. 

2. We construct a local deblurring dataset LODE with real-world captured locally blurred images and sharp ground truth. To our best knowledge, LODE is the first benchmark dataset for the local deblurring task. 

3. We propose a novel BladeNet for deep local deblurring. It consists of a data synthesis method, a blur perception module and a blur-guided spatial attention module, all of which are designed based on the characteristics of local blur. BladeNet is a flexible framework and combining it to current SotAs can further improve their performance.

4. We carry out extensive experiments for image deblurring. The results show BladeNet improves the PSNR by about $\mathbf{2.5}$ \textbf{dB} over the SotAs for local deblurring, while showing comparable performance for global deblurring.

\section{Related Works}

\subsection{Image Deblurring}
Traditional deblurring methods model the blurring process as the convolution between a latent sharp image and a blur kernel \cite{whyte2012non,gupta2010single} and try to retrieve both latent sharp image and blur kernel from the degraded blur image. These methods usually design hand-crafted image priors such as total variation \cite{chan1998total} or heavy-tailed gradient prior \cite{shan2008high} to regularize the inverse problem optimization. However, they are usually unrealistic and fail to capture the complex motion blur in real-world blurred images. 

Recently, with the rapid development of deep learning, kernel-free learning-based methods \cite{nah2017deep, kupyn2018deblurgan, kupyn2019deblurgan, gao2019dynamic, zhang2019deep, DBLP:conf/aaai/PurohitR20, suin2020spatially, yuan2020efficient, zhang2018dynamic, tao2018scale, lu2019unsupervised, zhang2020deblurring, zhang2019DPSR,kaufman2020deblurring} have been proposed. These methods use end-to-end deep networks to directly learn a mapping from a blurred image to a sharp one and achieve much superior performances. Nah \etal \cite{nah2017deep} propose a multi-scale CNN with adversarial loss in a coarse-to-fine manner. Gao \etal \cite{gao2019dynamic} propose a principled parameter selective sharing scheme. Zhang \etal \cite{zhang2019deep} propose a stacked multi-patch network for faster inference. Deformable convolution \cite{dai2017deformable} has also been introduced in \cite{DBLP:conf/aaai/PurohitR20,yuan2020efficient} to capture the spatial variant blur. Meanwhile, other deep architectures such as Recurrent neural networks (RNNs) and Generative Adversarial Networks (GAN) have also been used for image deblurring \cite{zhang2018dynamic,tao2018scale,kupyn2019deblurgan,lu2019unsupervised,zhang2020deblurring}. Among these methods, Zhang \etal \cite{zhang2018dynamic} formulate the deblurring process by an infinite impulse response (IIR) model and propose a novel end-to-end trainable spatially variant RNN for dynamic scene deblurring, while Tao \etal \cite{tao2018scale} propose a scale-recurrent network for deblurring task with less parameters. Other works use unsupervised methods and therefore do not rely on paired data. Kupyn \etal \cite{kupyn2018deblurgan} utilize a GAN \cite{goodfellow2014generative} network for deblurring with perceptual loss and further improve it with Feature Pyramid Network (FPN) \cite{kupyn2019deblurgan}. Lu \etal \cite{lu2019unsupervised} propose a CycleGAN-like unsupervised method for domain-specific deblurring based on disentangled representations with KL divergence loss. 

While the aforementioned methods have achieved remarkable progress, most deep learning based works only consider the global blur. Although few attempts have been \cite{schelten2014localized,pan2016soft} made to address the local motion blur, they are limited within the scope of localized kernel estimation, where deep learning based local deblurring still remains in blank.


\subsection{Deblurring Datasets}

Early deblurring datasets \cite{levin2009understanding,sun2015learning,chakrabarti2016neural} mainly consist of synthetic images with manually created blur kernel and are only used for evaluation. Levin \etal construct 32 test blur images using four latent sharp images and eight uniform blur kernels. Sun \etal \cite{sun2015learning} extend the blurred image set with sampled images from PASCAL VOC 2010 dataset \cite{everingham2010pascal} and 73 possible motion kernels. Meanwhile, Chakrabarti \etal \cite{chakrabarti2016neural} generate motion kernel by fitting spines on a grid and construct blurred image set with global kernel and sharp image patches from PASCAL VOC 2012 dataset \cite{pascal-voc-2012}. Gong \etal \cite{gong2017motion} generate blurred images with motion flow maps simulated from artificially created 3D motion trajectory and BSD500 dataset \cite{MartinFTM01}. Kupyn \etal  \cite{kupyn2018deblurgan} generate the blurred image with an image motion trajectory by Markov process. 
However, these datasets are unrealistic as they deviate far from the real-world scenarios.
In order to obtain realistic blurred images while also facilitate learning-based methods, large-scale deblurring datasets \cite{lai2016comparative,nah2017deep,Nah_2019_CVPR_Workshops_REDS} have been proposed using a high-speed camera recorded from realistic scenes.
Nah \etal \cite{nah2017deep} release the GOPRO dataset collected by GOPRO4 Hero Black camera which is composed of over 3000 images generated from 33 high frame rate videos. 
In order to provide blurred-sharp image pairs for deep model training, sharp frames captured by the high-speed camera are adopted as groundtruth and consecutive video frames are blended to generate the blurred images. Further, Nah \etal \cite{Nah_2019_CVPR_Workshops_REDS} release REDS dataset captured by GOPRO Hero6 Black camera which is composed of over 30000 blurred-sharp image pairs generated from 300 high frame rate videos. While these datasets greatly boost the development of deep learning based methods, they mostly focus on the global blur dominated by severe camera shake. The local deblurring problem, in terms of both dataset and archietecture, is largely neglected and lagging behind.

\section{Local Deblurring Dataset}

Data is the foundation of deep learning based approaches. Since there is no local deblurring dataset for training and evaluation, we propose the first real-world \textbf{lo}cal-\textbf{de}blur (LODE) dataset to facilitate relevant research. Different from previous datasets such as GOPRO or REDS where the images are dominated by global blur (which are mostly derived from deliberate camera shake), the proposed LODE focuses more on the local blur mostly caused by moving objects while the background is relatively sharp. LODE consists of $3,700$ images in 1920 $\times$ 1080 resolution recorded from various scenes. We briefly introduce the LODE dataset and more details can be found in the supplementary material.

\textbf{Recording}: We record over $3,700$ high-frame-rate (250 fps) video clips using a Sony ZV-1 camera under the program auto mode. Each video clip is 3 seconds long. We set the Sony SteadyShot mode on and record carefully to keep the background static and ensure each video clip contains at least one moving object.
Our background scenes include parks, zoos, aquariums, rivers, streets, libraries, playgrounds, shopping centers, etc. The moving objects include people, bicycles, cars, other vehicles, various animals and other items. 
The recording time covers day and night with different illumination conditions. 

\textbf{Blur synthesis}: Following previous works \cite{nah2017deep,Nah_2019_CVPR_Workshops_REDS}, we generate blurred images by averaging consecutive $M$ frames which are randomly sampled from the recorded video clips. We set the frame sequence length $M$ as an odd number randomly sampled from 7 to 13, then we set the mid-frame as the corresponding sharp latent image (ground truth). No compression or downsampling is made and the images have the same resolution as the recorded videos. We also calibrate the non-linear Camera Response Function (CRF) following \cite{nah2017deep}. Both video clips and the generated blurred-sharp image pairs will be released publicly.

\textbf{Partition}: LODE is composed of $2,700$ image pairs as the training set and $1,000$ image pairs as the validation set. Each image is generated from a different video. We make sure that there is no overlap of source videos between the training and validation sets.

\begin{figure*}
\begin{center}
\includegraphics[width= \linewidth]{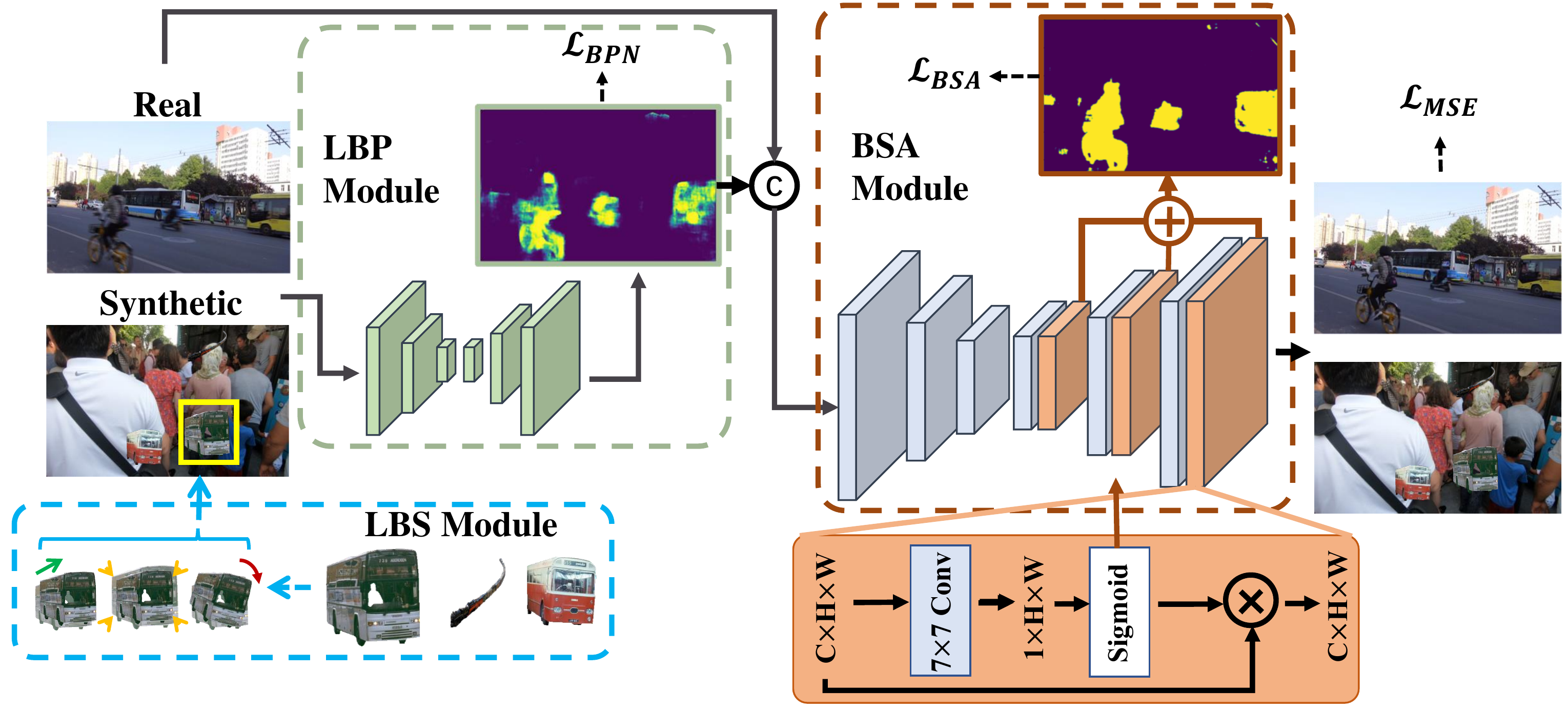}
\end{center}
\vspace{-5pt}
\caption{The proposed BladeNet for deep local deblurring. BladeNet has three main modules. The local blur synthesis (LBS) module utilizes an auxiliary dataset equipped with object segmentation masks to generate synthetic images with local blur. In this way, the model can learn the pattern of local blur from data. The local blur perception (LBP) module captures the blur region and predicts the probability map of the blurred area, such that local blur areas could be restored and sharp background is retained. The blur-guided spatial attention (BSA) module uses the segmentation mask trained in the data synthesis module as a supervision signal to guide the spatial-attention to focus on those locally blurred areas. They are all designed based on the characteristics of the local blur.}
\vspace{-5pt}
\label{fig:blade_arch}
\end{figure*}

\section{Blur-aware Deblurring Network}

Given a locally blurred image, we wish the deep network can effectively perceive the blurred region and pay more attention to those local regions. To achieve this goal, we propose the Blur-aware Deblurring Network (BladeNet). BladeNet is mainly composed of a Local Blur Synthesis (LBS) module for locally blurred training image generation, a Local Blur Perception (LBP) module for blur region prediction, and a main deblurring network equipped with Blur-guided Spatial Attention (BSA) module for localized deblurring. The whole BladeNet architecture is illustrated in Figure \ref{fig:blade_arch}.

\subsection{Local Blur Synthesis Module}

We firstly propose a simple yet generalizable Local Blur Synthesis (LBS) method that can effectively generate synthetic local blurred images for training. Unlike previous blurred image synthesis methods that generate blurred images using synthetic blur kernels, we focus on the local blur generation and we resort to an auxiliary dataset with high-quality segmentation annotation. Such datasets can be easily found in segmentation tasks such as PASCAL VOC~\cite{everingham2010pascal, pascal-voc-2012} and MS COCO~\cite{lin2014microsoft}. With the segmentation mask, we can crop various objects from the auxiliary dataset and create corresponding movements to generate local motion blur. In practice, we choose PASCAL VOC 2012 as an auxiliary dataset and REDS as a background image dataset.

For each synthetic image with local blur, the Local Blur Synthesis module takes one sharp image from REDS as background and two cropped images from PASCAL VOC 2012 as moving objects. With the pixel-level segmentation annotation, the cropped images will not introduce any irrelevant visual content. Due to the mismatch in resolution between REDS and PASCAL VOC 2012, we rescale the cropped object patches by 1.2 so that they will not be too small. Then we follow a similar blurred image generation process described in Section 3. To be specific, we randomly choose an initial position for each object on the background image. Then we simulate the movement of these objects through 3 different transformations. Assume that the objects move in a 3-dimensional space formed by the plane of image (x and y axis) and axis perpendicular to the image plane (z axis). We use translation of the objects on image to simulate the movement along x and y axis, use scaling to simulate the movement of objects along z axis and use rotation to simulate the rotation of the object. For synthetic local blur generation, we first create a trajectory with length $L$ where $L$ is randomly selected as an odd number ranging from 7 to 13. At each step in the trajectory, the three pre-defined transformations and their order are randomly selected to create a synthetic frame. Finally, the locally blurred synthetic images $I_{LBS}$ is generated by averaging the synthetic frames along the trajectory, and the middle sharp synthetic frame is taken as ground-truth latent image $I_{Sharp}$. We also keep the synthetic object segmentation mask and denote as $M_{seg}$.

In order to enhance the generalizability of the LBS module and avoid overfitting, we also paste some static cropped objects to the background at fixed position. In this way, the model trained on the synthetic data will be forced to focus on those locally blurred objects, rather than any object that stands out of the background content. It is also worth noting that the proposed LBS Module holds several advantages. Firstly, since both sharp background images and object images with segmentation annotation can be easily acquired, the LBS is free of any extra cost. Secondly, the LBS is a universal and generalizable data synthesis method. The combination of diverse backgrounds and various objects with different motion enables the LBS to create as many synthetic images as possible, and largely improves the generalizability of the deep networks trained on them. Finally, since the segmentation annotation can be acquired along with the moving objects, it can provide extra information for deep networks training. We will show that fully exploiting such information with specially designed modules will be enormously beneficial for deep networks for local blur perception.

\subsection{Local Blur Perception Module}

According to the characteristics of local blur, we propose the LBP module to automatically perceive and localize the blurred region. The LBP module has a structure of a U-Net. It takes a locally blurred image $I_{B} \in R^{C\times H\times W}$ and predicts a probability map $M_{LBP} \in R^{2\times H \times W}$ indicating whether a pixel belongs to a locally blurred region or not. Then the probability map from LBP is concatenated with the original blurred image $I_B$ and sent to the main deblurring network. Since we have the pixel-level segmentation mask $M_{seg}$ of the synthetic blurred objects generated from the LBS Module, we train the LBP Module with pixel-level supervision and a pixel-wise Binary Cross Entropy loss is adopted as follows,
\begin{equation}
L_{LBP} = CrossEntropy(M_{LBP}, M_{seg})
\end{equation}

In practice, we first train the LBP module independently to equip it with the blur perception capability. Then we finetune the LBP module jointly with the main deblurring network so that local blur information will be better captured and improve the local deblurring performance.

\subsection{Blur-guided Spatial Attention Module}

Apart from training an independent LBP module to provide possible locally blurred regions, we also design a Blur-guided Spatial Attention (BSA) module to guide the main deblurring network to focus on those locally blurred regions. Specifically, the main deblurring network also has a structure of a U-Net. Given a feature map $F \in R^{C\times H \times W}$, the spatial attention first generates the attention map as follows
\begin{equation}
 M_{SA} = \mathrm{Sigmoid}(\mathrm{Conv}_{7\times 7}(F)) \quad M_{SA} \in R^{1\times H \times W}
\end{equation}
Then the attention map attends to the original feature map so that important regions, which are locally blurred regions in our case, are highlighted:
\begin{equation}
F_{Att} = F \odot M_{SA}
\end{equation}
where $\odot$ denotes element-wise multiplication. While ordinary attention module learns the important regions implicitly, thanks to the segmentation mask produced by the LBS module, we are able to explicitly supervise the spatial attention to focus on those locally blurred regions. In practice, we keep the first three downsampling stages in the U-Net unchanged to maintain the capability of global deblurring. Then we add blur-guided spatial attention in the last three upsampling stages to highlight local deblurring. If we denote $M_{SA}^i \in R^{1 \times H_i \times W_i}$ as the attention map at $i^{th}$ layer, where $H_i, W_i$ are the feature map's shape, we first upsample the feature map to the original input spatial size $[H_0, W_0]$, then we add them together and supervise them with the segmentation mask:
\begin{equation}
L_{BSA} = L_{DICE}( \sum_i \mathrm{upsample}(M^i_{sa}), M_{seg})
\end{equation}
where $L_{DICE} = 1 - \frac{2\left | \sum_i M^i_{sa} \cap M_{seg} \right |}{ \left | \sum_i M^i_{sa} \right | + \left | M_{seg} \right | }$ denotes the DICE loss \cite{milletari2016v}. More network details can be found in the supplementary material.

\begin{table*}[t]
\caption{Comparison of different deblurring datasets.}
\vspace{0pt}
\begin{center}
\begin{tabular}{c|c|c|c|c|c}
\hline
 Dataset & Levin \etal \cite{levin2009understanding} & Lai \etal \cite{lai2016comparative} & GOPRO \cite{nah2017deep}  & REDS \cite{Nah_2019_CVPR_Workshops_REDS} & LODE\\ 
\hline \hline
 Synthetic/Real & Synthetic & Real & Real & Real & Real \\
 Blur Type & global & global & global &  global & local \\
 Lantent Images & 4 & 100 & 3214 & 30000 & 3700 \\ 
 Blurred Images & 32 & 100 & 3214 & 30000 & 3700 \\
 Kernels / Trajectories & 8 & 100 & Camera & Camera & Camera \\
 Frame Per Second & - & - & 240 & 120 (virtual 1920) & 250 \\
 Scenes & 4 & 100 & 33 & 300 & $\geq$ 50 \\
 Resolution & $255\times255$ & $\leq 900\times1200$ & $720\times1280$ & $720\times1280$ & $1080\times1920$ \\
 \hline 
\end{tabular}
\end{center}
\label{tab:datacompare_table}
\end{table*}

\begin{figure}[t]
\begin{center}
\includegraphics[width= 0.8\linewidth]{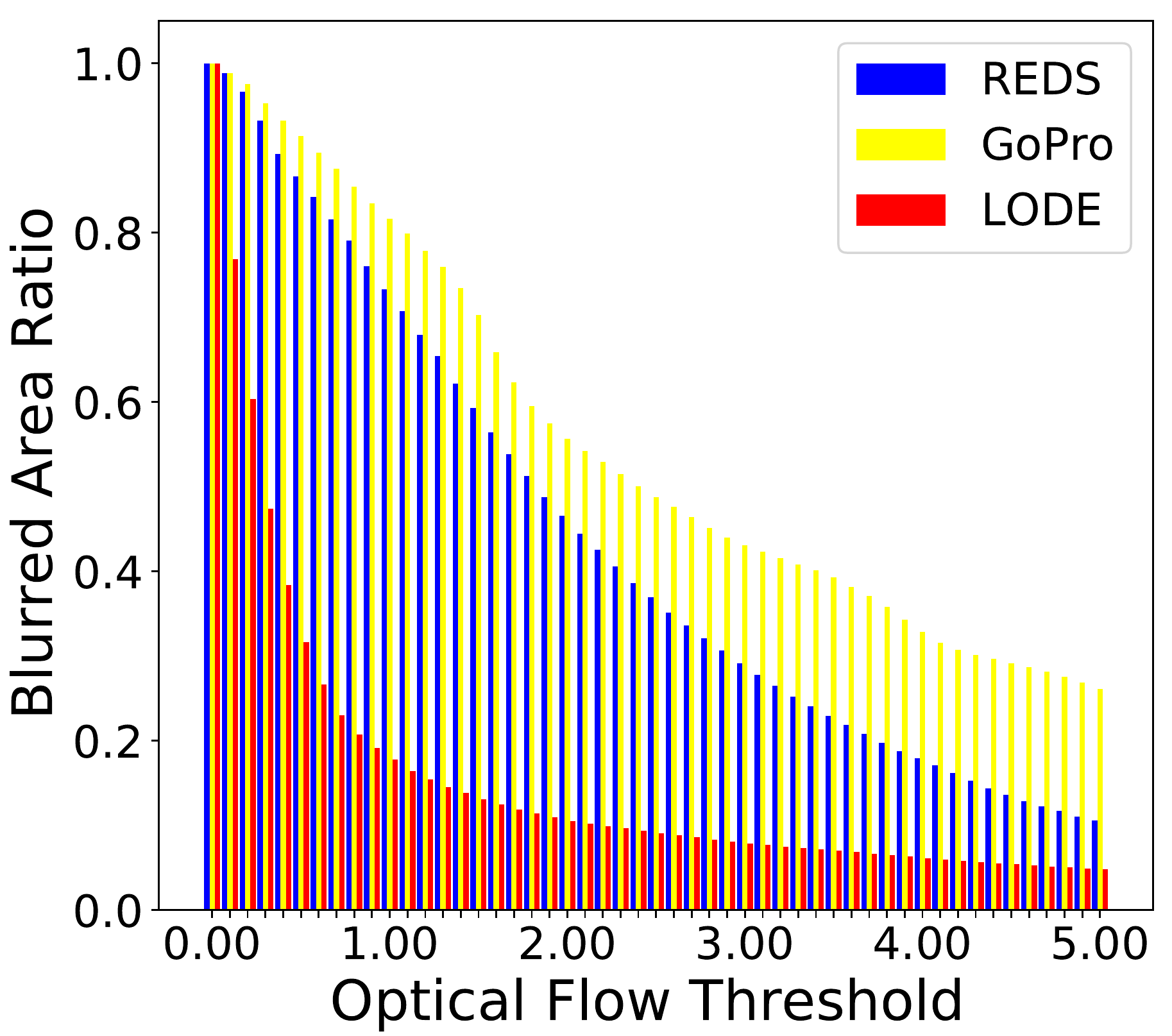}
\end{center}
\vspace{-10pt}
\caption{Blurred area ratio with respect to optical flow threshold from $0$ to $5$ with $step=0.1$. It shows that LODE contains less large areas of global motion blur than GOPRO and REDS.}
\vspace{-10pt}
\label{fig:area_ratio}
\end{figure}

\begin{table*}
\caption{Main results on LODE dataset. All comparison results are generated using the publicly available codes.}
\vspace{-5pt}
\begin{center}
\begin{tabular}{c|c|c|c|c|c|c|c}
\hline
Method  &  SRN-Deblur \cite{tao2018scale} & DPSR \cite{zhang2019DPSR} & SDNet \cite{zhang2019deep} & DeblurGAN-V2 \cite{kupyn2019deblurgan} & BladeNet$^-$ & BladeNet &  BladeNet$^+$  \\
\hline
\hline
PSNR & 30.71 & 31.84 & 31.41 & 27.29 & 31.89 & 34.12 & 34.36 \\
SSIM &   0.9219 & 0.9241 & 0.9321  & 0.8691 & 0.9382 & 0.9467 & 0.9487 \\ \hline
 
\end{tabular}
\end{center}
\label{tab:sota_compare}
\end{table*}

\begin{figure*}[!h]
\begin{center}
\includegraphics[width= \linewidth]{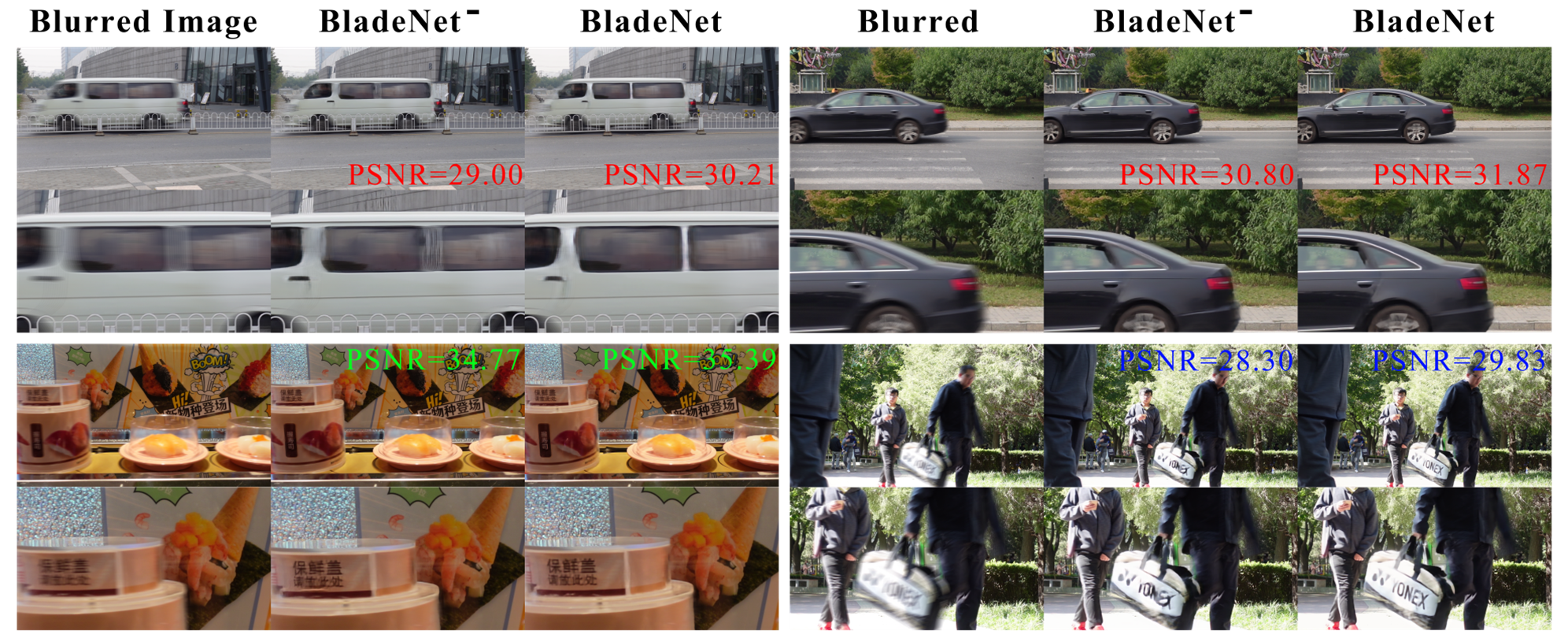}
\end{center}
\vspace{-10pt}
  \caption{Visual comparison of BladeNet on Local Deblurring. The results show that the proposed three modules can greatly recover the blurred details.  (e.g. `YONEX' letters in the last row.)}
\vspace{-5pt}
\label{fig:blade_result}
\end{figure*}

\subsection{Training Strategy}
In practice, we first train the LBP module independently. Then we train the BSA-equipped main deblurring network and the LBP module alternatively, that is, we first fix the LBP module and train the main deblurring network with mean square error (MSE) loss $L_{MSE}$, then we train the entire BladeNet using the synthetic data generated by LBS module with the following loss,
\begin{equation}
    L =  \lambda_{1} L_{MSE} + \lambda_{2} L_{LBP} + \lambda_{3} L_{BSA}
\end{equation}

\section{Experiments}

We first evaluate and analyze the differences between previous global deblurring datasets and the proposed LODE dataset. Then we conduct extensive experiments to show that the proposed BladeNet is able to significantly outperform previous methods and largely improve the local deblurring performance.

\begin{table*}[ht]
\caption{Performance comparison between the original REDS dataset and REDS+VOC generated by LBS module.}
\begin{center}
\begin{tabular}{c|c|c|c|c|c|c|c|c}
\hline
 & \multicolumn{4}{c|}{Evaluation on LODE} & \multicolumn{4}{c}{Evaluation on REDS} \\
\hline 
 \multirow{2}{*}{Method} & \multicolumn{2}{c|}{REDS} & \multicolumn{2}{c|}{REDS+VOC} & \multicolumn{2}{c|}{REDS} & \multicolumn{2}{c}{REDS+VOC} \\ \cline{2-9} 
 & PSNR & SSIM & PSNR & SSIM & PSNR & SSIM & PSNR & SSIM \\
\hline
\hline
 SRN \cite{tao2018scale} & 30.71 & 0.9219 & 31.67     & 0.9249      & 31.26 & 0.8922 & 30.74    & 0.8771      \\
 DPSR \cite{zhang2019DPSR} & 31.84 & 0.9241 & 31.98 & 0.9242 & 25.72 & 0.7531 & 25.76 & 0.7546\\
 SDNet \cite{zhang2019deep} & 31.41 & 0.9321 & 33.18 & 0.9415 & 31.34 & 0.8928 & 30.20 & 0.8722      \\
 DeblurGan-V2 \cite{kupyn2019deblurgan} & 27.29 & 0.8691 & 29.48 & 0.9033 & 26.66 & 0.7823 & 26.31& 0.7704 \\ 
 BladeNet$^-$        & 31.89 & 0.9382 & 33.63 & 0.9463 & 30.40 & 0.8737 & 30.59 & 0.8741 \\ \hline 
\end{tabular}
\end{center}
\label{tab:aug_result}
\end{table*}

\begin{figure*}[ht]
\begin{center}
\includegraphics[width= \linewidth]{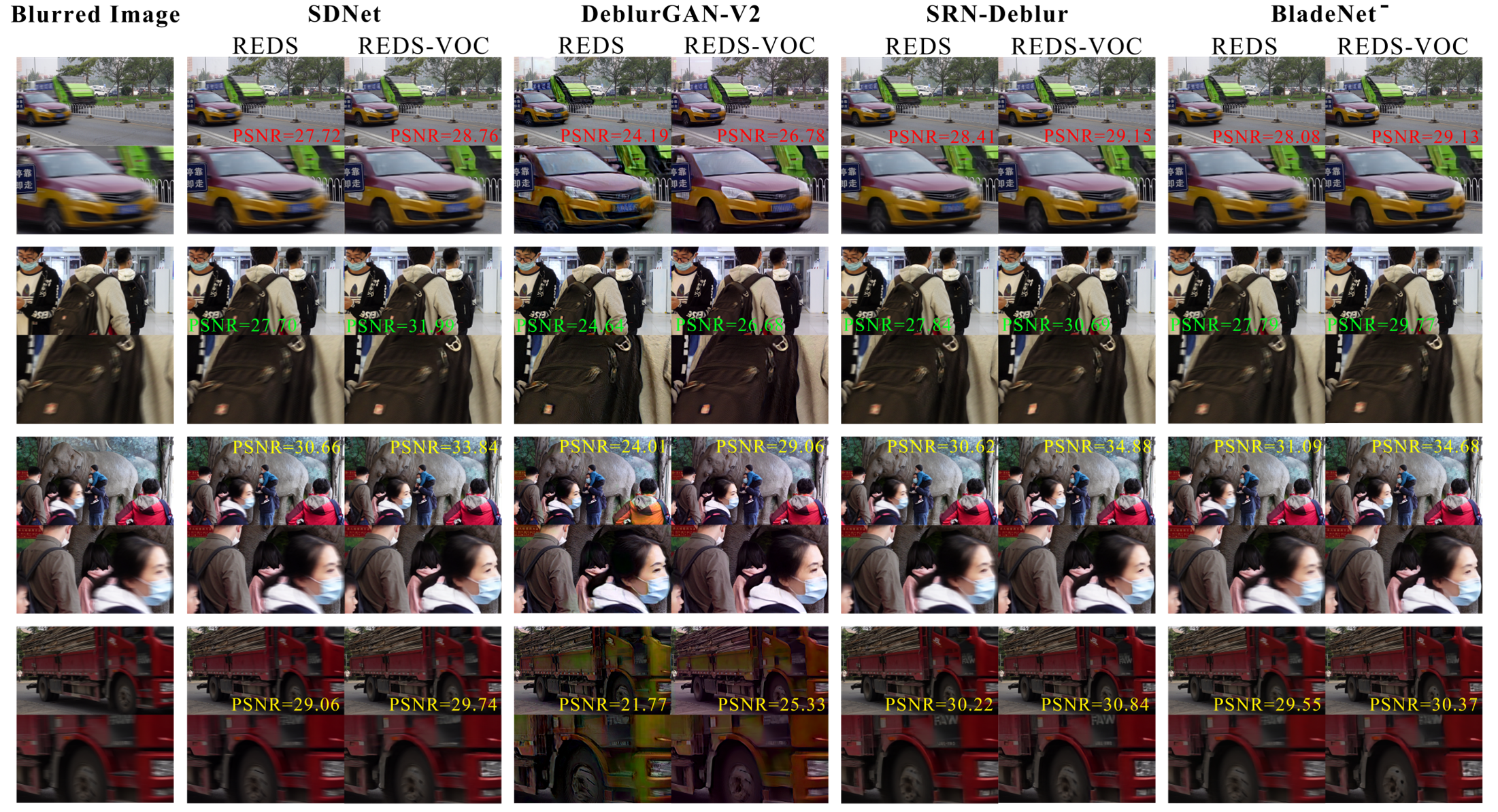}
\end{center}
\vspace{-5pt}
  \caption{Visual comparisons of with LBP module (REDS+VOC) and without LBP module (REDS).}
\vspace{-5pt}
\label{fig:aug_vis}
\end{figure*}

\subsection{Statistical Analysis of LODE Dataset}

The main feature of the proposed LODE dataset is the local blur that mostly derives from object motion. In order to quantify this characteristic statistically, we compare the blurred region area between previous global deblurring datasets \cite{nah2017deep,Nah_2019_CVPR_Workshops_REDS} and the LODE dataset. Intuitively, globally blurred images are dominated by the global camera shake, thus the blur area covers a large portion of the whole image. On the other hand, the locally blurred images are mostly degraded by the local object motion, thus only a small fraction of pixels are involved. In practice, we first define blurred area ratio as $\frac{area\_of\_blurred\_region}{area\_of\_the\_image}$ where the blurred region is defined as pixels whose optical flow norm is larger than a certain threshold. Then we can plot the blurred area ratio with respect to the optical flow threshold and the result is shown in Figure \ref{fig:area_ratio}. For each dataset, we randomly select a sequence of 2 frames on each 120 fps REDS video and a sequence of 3 frames on each 240 fps GOPRO video and 250 fps LODE video to avoid the impact from frame rate differences. We use Deepflow in OpenCV to calculate optical flow.  As shown in Figure \ref{fig:area_ratio}, the global deblurring datasets such as GOPRO and REDS have large blurred area ratio even when the threshold is high, indicating that blur caused by large global motion (i.e. severe camera shake) is dominating. On the other hand, the curve of LODE dataset shows a long-tailed distribution, indicating that only a small fraction of large motion pixels (i.e. moving objects) exists. This result distinguishes our LODE dataset and the local deblurring problem from previous studies. Furthermore, we list the statistics of previous datasets and the LODE dataset for a detailed comparison as shown in Table \ref{tab:datacompare_table}. We show that our LODE dataset is realistic, large-scale, high-resolution and diverse, and we believe it will further facilitate related research on the local deblurring problem.

\subsection{BladeNet Performance}

\textbf{Implementation detail.} We train the BladeNet for 400 epochs with Adam optimizer and initial learning rate 0.001 decayed by 0.5 every 100 epochs. 
The hyperparameters $\lambda_{1}, \lambda_{2}, \lambda_{3}$ are set to 1.0, 0.25, 0.01 respectively. For baseline methods, we follow their origin settings and use their public available codes to produce corresponding results.
More details can be found in the supplementary material.

\textbf{Comparison with state-of-the-art methods.} We compare the proposed BladeNet with state-of-the-art methods on the LODE dataset to investigate their performances on the local deblurring problem. The BladeNet$^-$, BladeNet, BladeNet$^+$ denote the proposed BladeNet without LBS, LBP and BSA modules, the standard BladeNet, and the BladeNet further finetuned on LODE, respectively. The result in Table \ref{tab:sota_compare} shows that the proposed BladeNet is able to outperform previous SotA methods by a large margin (2.5 dB in PSNR) on the local deblurring problem even without accessing to any locally blurred training data. Moreover, finetuning on LODE will further improve the local deblurring performance. We also qualitatively compare the original blurred image, BladeNet$^-$'s result and BladeNet's result as illustrated in Figure \ref{fig:blade_result}. The results show that compared to the BladeNet$^{-}$ whose outputs are still blurred and may contain artifacts (e.g. distortion of letters `YONEX' in the last row), the standard BladeNet is able to recover sharper images with more natural details.

\begin{table}
\caption{Ablation study for LBP and BSA modules}
\begin{center}
\begin{tabular}{c|c|c}
\hline
Method & PSNR & SSIM \\
\hline
\hline
w/o LBP/BSA  & 33.63  & 0.9463  \\
LBP Only & 33.88  & 0.9432  \\
BSA Only & 33.92  & 0.9458   \\
LBP+BSA & 34.12  & 0.9467  \\ \hline

\end{tabular}
\end{center}
\vspace{-10pt}
\label{tab:module_result}
\end{table}

\begin{table}
\caption{Result of LODE finetuning.}
\vspace{-10pt}
\begin{center}
\begin{tabular}{c|c|c|c|c}
\hline
\multirow{3}{*}{Methods} & \multicolumn{2}{c|}{Train on} & \multicolumn{2}{c}{Finetune on} \\ 
 & \multicolumn{2}{c|}{REDS+VOC} & \multicolumn{2}{c}{LODE} \\ \cline{2-5} 
 & PSNR & SSIM & PSNR & SSIM \\
\hline
\hline
SRN \cite{tao2018scale} & 31.67 & 0.9249 & 33.18 & 0.9338 \\
DPSR \cite{zhang2019DPSR} & 31.98 & 0.9242 & 32.35 & 0.9301 \\
SDNet \cite{zhang2019deep} & 33.18 & 0.9415 & 33.67 & 0.9394 \\
DeblurGan-V2 \cite{kupyn2019deblurgan} & 29.48 & 0.9033 & 31.28 & 0.9228 \\ \hline 
BladeNet (Ours)   & 34.12  & 0.9467 & 34.36 & 0.9487  \\ \hline 
\end{tabular}
\end{center}
\label{tab:finetune_result}
\vspace{-5pt}
\end{table}

\textbf{Effectiveness of Local Blur Systhesis}
In order to prove the effectiveness and generalizability of the LBS module on local deblurring problem, we show by experiments that LBS module can not only greatly promote the performance, but is also a universal strategy that can be easily combined with other SotA methods. The result in Table \ref{tab:aug_result} demonstrates that the LBS module (denoted as REDS+VOC) is able to bring significant improvements (a largest of 2.2dB in PSNR) on the local deblurring task while it is also beneficial for global deblurring. Meanwhile, since LBS requires no locally blurred data for training, it brings a free lunch for both local and global deblurring. Finally, we provide the qualitative comparison in Figure \ref{tab:aug_result}. The results show that previous SotAs often fail to restore the latent sharp images from the locally blurred ones. By contrast, the LBP-produced training data largely improves the visual quality, especially on common moving objects like cars (1st row) and persons (3rd row).



\begin{figure}
\begin{center}
\includegraphics[width= \linewidth]{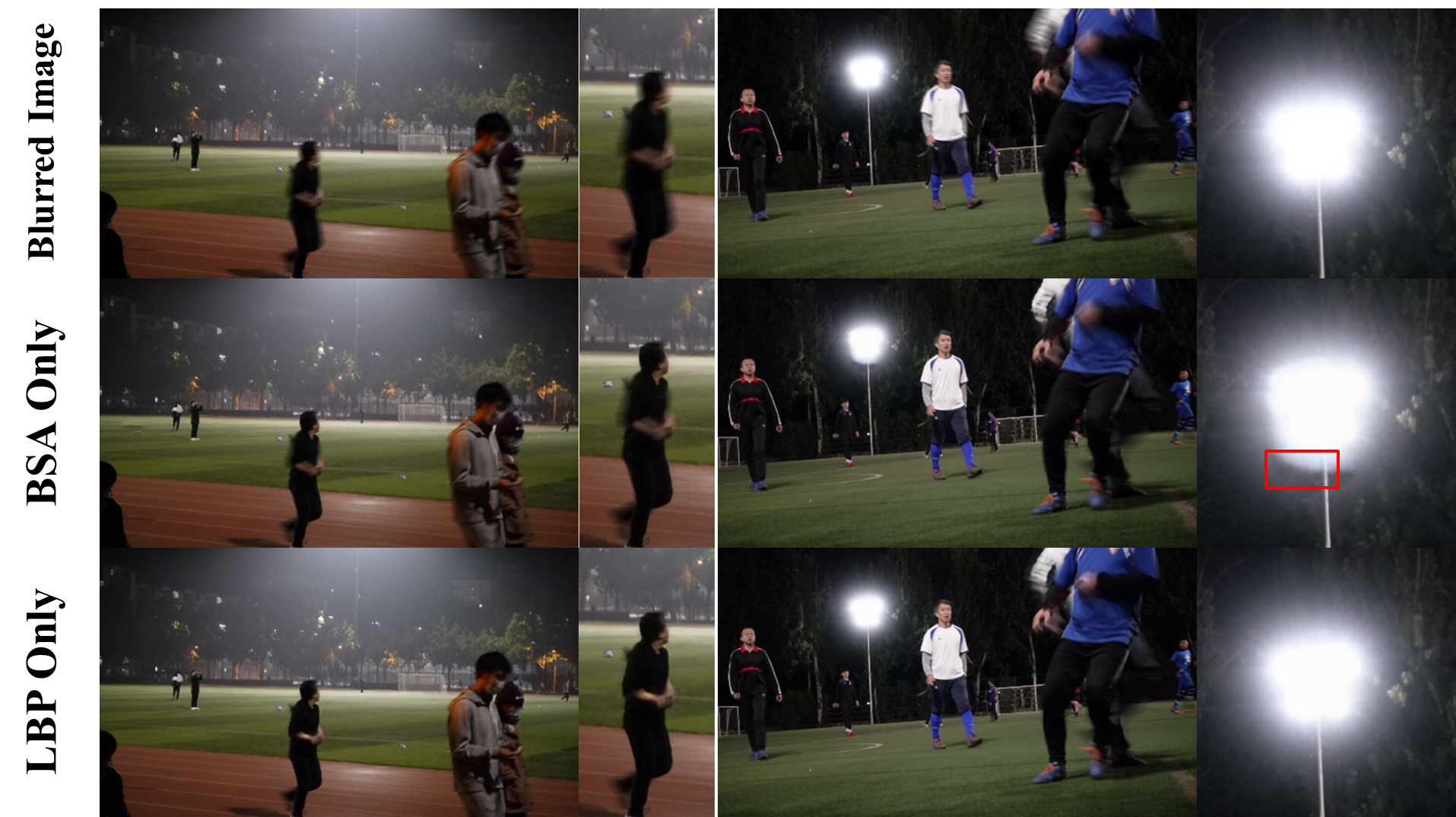}
\end{center}
\vspace{-10pt}
  \caption{Nighttime deblurring scenario.}
\vspace{-10pt}
\label{fig:nighttime_vis}
\end{figure}

\textbf{Effectiveness of Local Blur Perception and Blur-guided Spatial Attention.} To verify the effectiveness of the proposed LBP and BSA modules, we conduct component analysis as illustrated in Table \ref{tab:module_result}. The results show that both LBP and BSA contribute a large portion to the total improvements. Moreover, we provide the LBP and BSA's output in Figure \ref{fig:blade_arch} and show that both LBP and BSA are able to localize those blurred moving objects effectively and accurately. Note that both LBP and BSA are only trained with synthetic data generated by LBS, but they generalize well on realistic scenes, as they can effectively neglect those static objects and focus on those moving blurred objects. Finally, we investigate a common yet challenging scenario: nightime deblurring as illustrated in Figure \ref{fig:nighttime_vis}. The results show that LBP can protect the original images from being over-deblurred.

\begin{figure}
\begin{center}
\includegraphics[width= \linewidth]{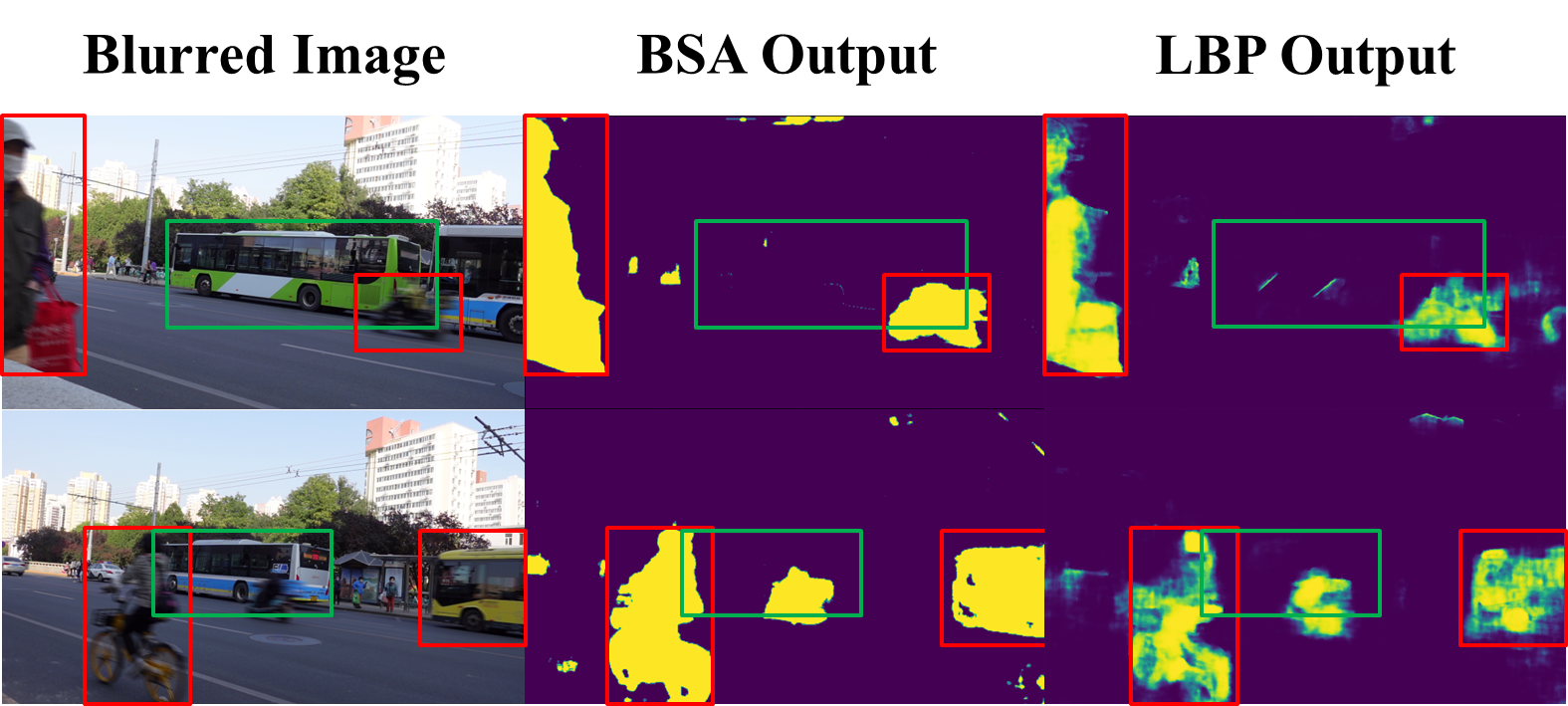}
\end{center}
\vspace{-10pt}
  \caption{Visualization of BSA and LBP's outputs. Both BSA and LBP can effectively attend to those blurred moving objects (red boxes) and ignore sharp static objects (green boxes).}
\vspace{-5pt}
\label{fig:sa_mask_vis}
\end{figure}

\textbf{Finetuning on LODE.} We further finetune various methods on LODE dataset. Table \ref{tab:finetune_result} shows that while LBS-generated synthetic data is beneficial for local deblurring, finetuning on LODE training set can further improve the local deblurring performance, which is also in accordance with our intuition. 

\vspace{-5pt}
\section{Conclusion}
In this paper, we raise the attention to the local deblurring problem and demonstrate that local deblurring is intrinsically different from global deblurring investigated by previous works. To tackle the local deblurring problem, we introduce the first real-world local deblurring dataset LODE. We further propose BladeNet which is designed based on the properties of local deblur. The BladeNet is composed of three components: the Local Blur Synthesis Module which generates synthetic locally blurred image, the Local Blur Perception Module which detects and localizes the blur region and the Blur-guided Spatial Attention Module which highlights the blur region for the deblurring network. We analyze the LODE's characteristics on the local deblurring problem and demonstrate by extensive experiments the superiority of the proposed BladeNet. We believe the proposed LODE dataset and the BladeNet framework will shed light on the local deblurring problem and further facilitate related research on this topic.

{\small
\bibliographystyle{ieee_fullname}
\bibliography{egbib}
}

\clearpage

\renewcommand\thesection{\Alph{section}}
\setcounter{section}{0}

\section{LODE Dataset}
We first introduce the calibration details for LODE dataset generation. Then we present image samples in the LODE dataset to give a intuitive comprehension of local blur.
\subsection{Calibration} 
When averaging sharp frames captured by high-speed camera to generate blurred images, calibration is a necessary procedure. The reason that we could average sharp frames for blurred image generation is that during the exposure, the blurred sensor signal can be seen as an integral of sharp sensor signals over time. Then the sensor signal is transformed into RGB pixel values by a non-linear Camera Response Function (CRF). The whole process can be expressed as follows: 
\begin{align}
    B = g \left ( \frac{1}{T} \int_{t=0}^{T} S(t)  dt \right ) \approx g \left ( \frac{1}{M} \sum_{i=0}^{M-1} S[i] \right )
\end{align}
Where $B$ denotes the synthetic blurred frame. $T,M$ denotes the exposure time and the number of sampled frames respectively. $S(t),S[i]$ denotes latent frame at time $t$ and $i$-th sampled latent frame in the sensor space respectively. Then the CRF $g$ transforms $S[i]$ into RGB pixel values $\hat{S}[i] = g(S[i])$.


Since the non-linear CRF is unknown to us, we estimate the nonlinear CRF by a common practice \cite{nah2017deep}:
\begin{align}
    g(x) = x^{1 / \gamma}
\end{align}
where $\gamma$ is set to 2.2. With the estimated CRF, we convert the observed RGB pixel values $\hat{S}[i]$ into sensor signal space $S[i]=g^{-1}(\hat(S)[i])$ and generate the blurred sensor signal by averaging the latent sharp ones $S[i]$. Finally the blurred RGB image is obtained by converting the blurred sensor signal back to the RGB space.

\subsection{Samples in the LODE dataset} 
\begin{figure*}
\begin{center}
\includegraphics[width= \linewidth]{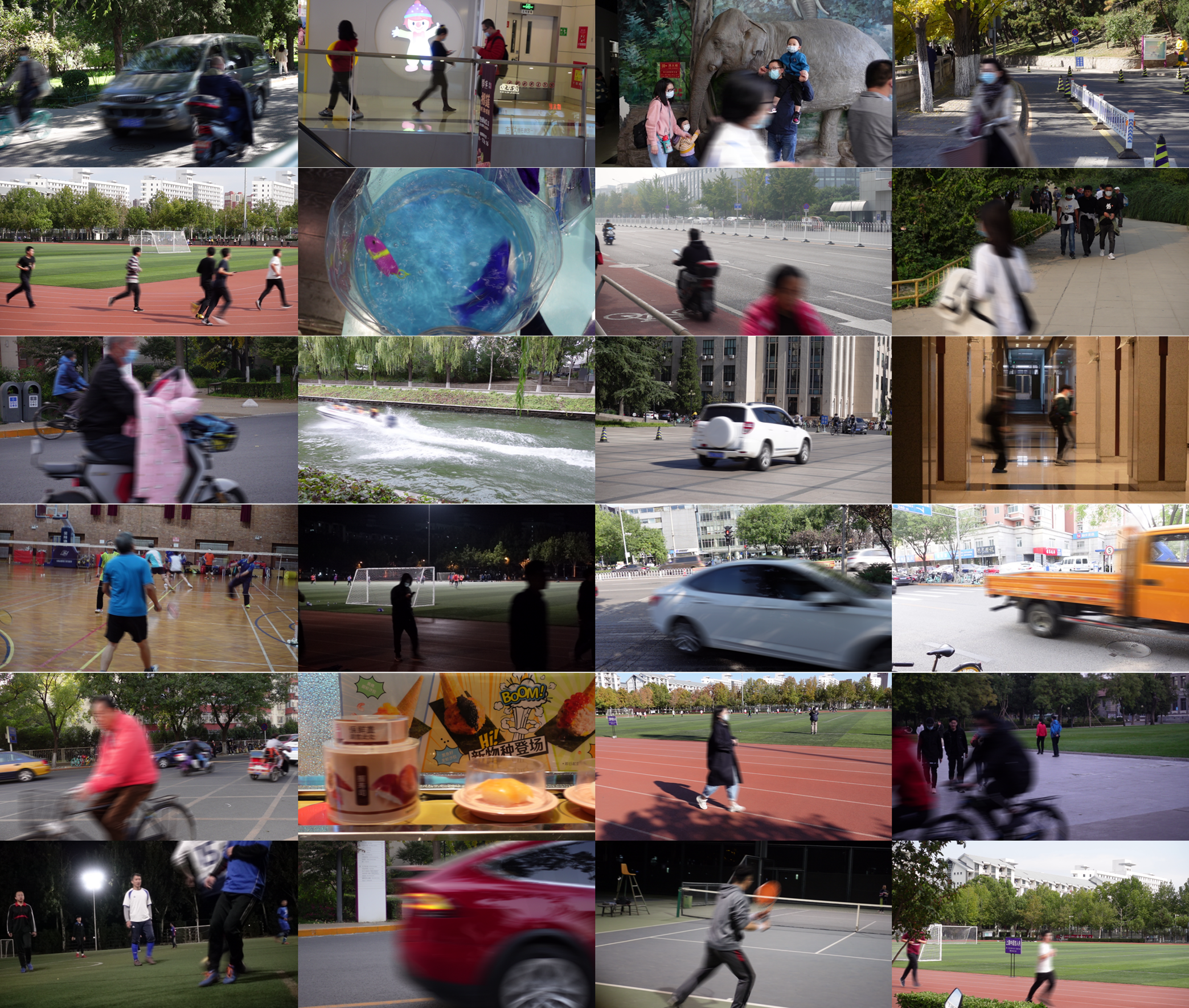}
\end{center}
   \caption{Sample images from LODE dataset.}
\label{fig:LODE_sample}
\end{figure*}

We present several samples in LODE dataset from various scenes shown in Fig \ref{fig:LODE_sample}. We show that the proposed LODE dataset is diverse and realistic. Most of the samples are from the daily life scenes, which makes it practical for real-world applications. It also includes various scenes from daytime to nighttime, from outdoor to indoor, so that it can fully investigate the generlizability for different  deblurring algorithms.

\begin{figure*}
\begin{center}
\includegraphics[width= \linewidth]{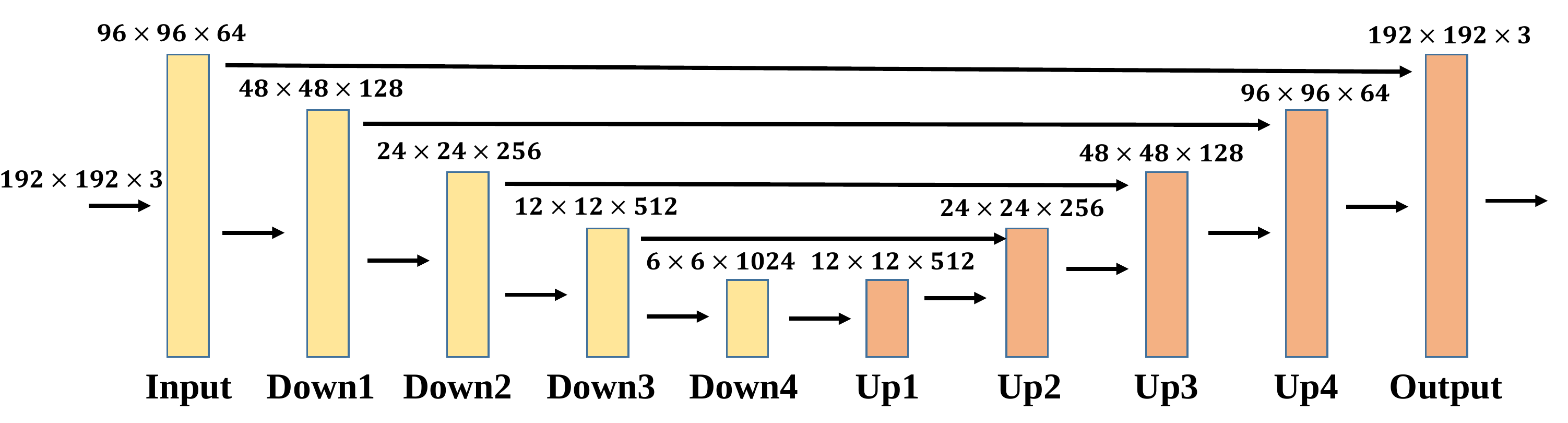}
\end{center}
   \caption{Detailed Structure of BladeNet.}
\label{fig:illustration_supp}
\end{figure*}

\begin{figure*}
\begin{center}
\includegraphics[width= \linewidth]{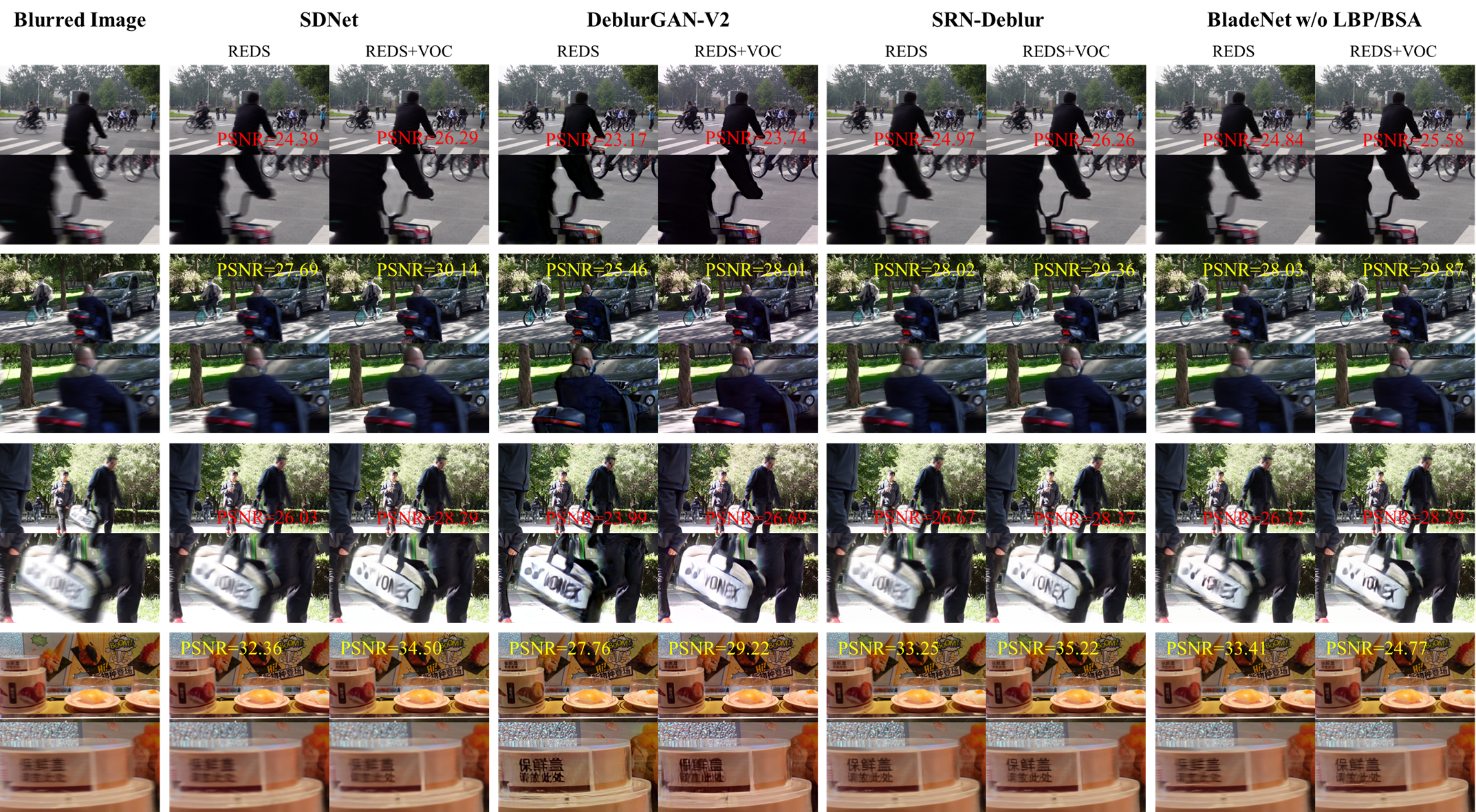}
\end{center}
   \caption{Visual comparison of with LBS module (REDS+VOC) and without LBS module (REDS).}
\label{fig:LBP_compare_sup}
\end{figure*}

\begin{figure*}
\begin{center}
\includegraphics[width= \linewidth]{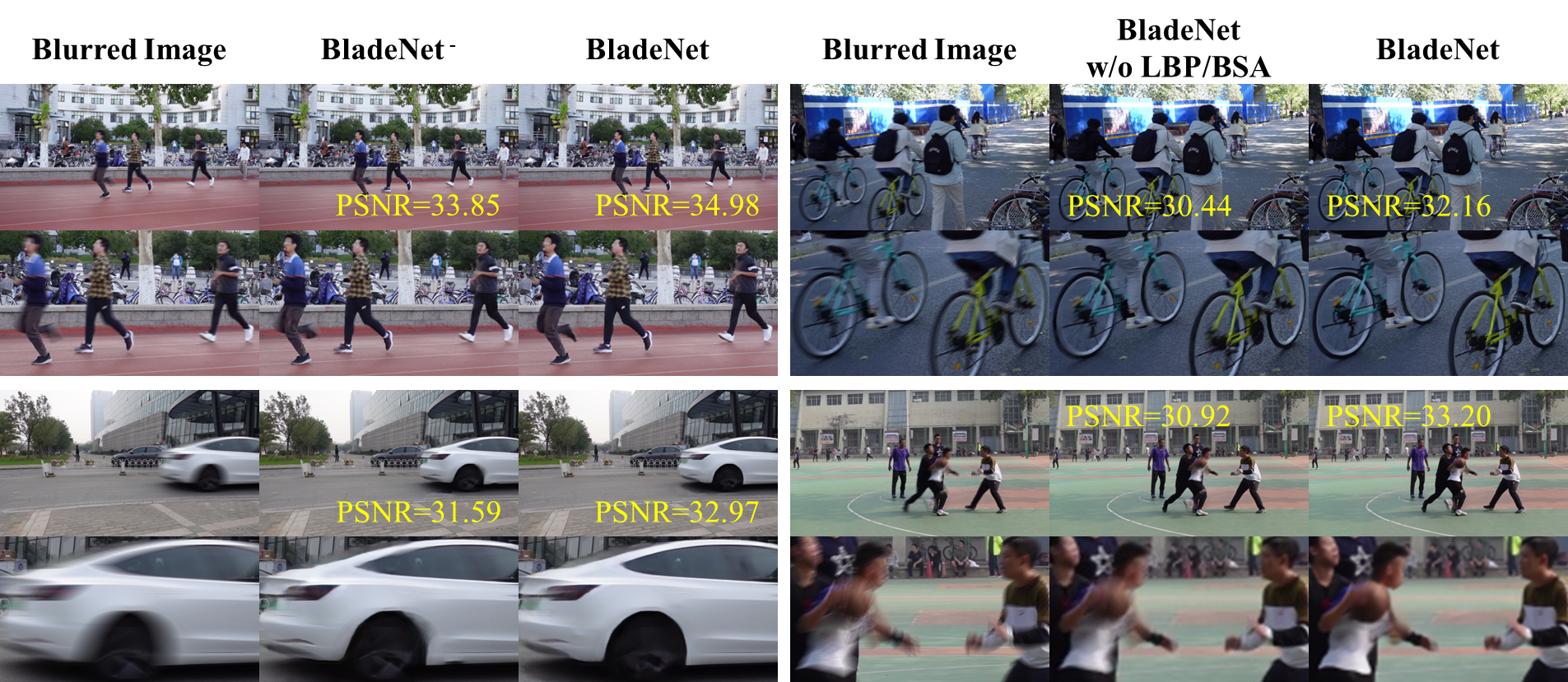}
\end{center}
   \caption{Visual comparison of BladeNet with and without the LBP/BSA modules.}
\label{fig:LBP_BSA_sup}
\end{figure*}

\begin{figure*}
\begin{center}
\includegraphics[width= \linewidth]{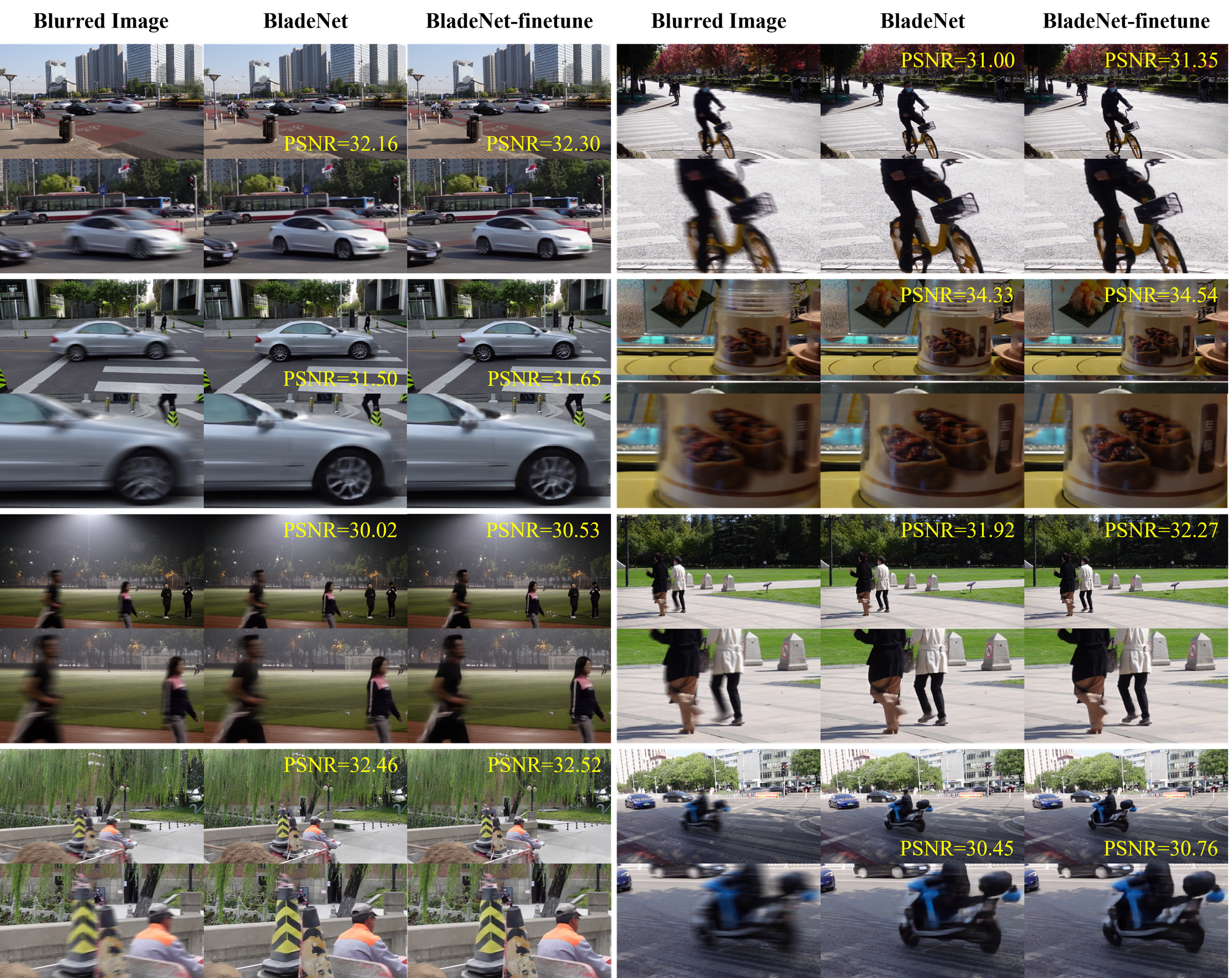}
\end{center}
   \caption{Result of LODE finetuning.}
\label{fig:finetuned_result_sup}
\end{figure*}

\section{BladeNet}
\subsection{Detailed Network Structure}
As mentioned before, the LBP module and the main deblurring network in BladeNet both has a U-Net structure. The detailed network architectures can be found in Fig \ref{fig:illustration_supp}. The U-Net adopted here has four downsampling blocks (Down1, Down2, Down3, Down4) and four upsampling blocks (Up1, Up2, Up3, Up4). The downsampling block structure is `Maxpool-Conv-BN-ReLU-Conv-BN-ReLU'. The upsampling block structure is `bilinearUpsample-Conv-BN-ReLU-Conv-BN-ReLU-Concat'. For adding BSA module, we replace the block structure to `bilinearUpsample-Conv-BN-ReLU-Conv-BN-ReLU-BSA-Conv-BN-ReLU-Conv-BN-ReLU-Concat' in Up2, Up3 and Up4.

It is worth noting that BladeNet is not only effective on local deblurring, but is also computational efficient. It has 40.8M parameters, which is comparable or even less compared to state-of-the-art methods (SRN(33.6M), SDNet(86.8M)), with much superior local deblurring performance.

\subsection{Implementation Detail}
All experiments are implemented using PyTorch 1.0.1 and conducted on a server with Intel(R) Xeon(R) E5-2660 CPU and 8 NVIDIA GeForce 2080 Ti GPUs. 

For BladeNet$^-$ (BladeNet without the LBS, LBP and BSA modules, only the main deblurring network) training, we use REDS \cite{Nah_2019_CVPR_Workshops_REDS} as training data and mean square error (MSE) loss to train the network. For BladeNet training, we use REDS combined with the REDS+VOC synthetic data generated by LBS module as training data. We first train the LBP module for 100 epochs. An alternating training strategy is adopted where each training epoch is equally divided into 2 stages: 1) we use REDS images as training data and adopt MSE loss in the first stage. 2) We use the synthetic REDS+VOC images along with their segmentation masks generated by the LBS module as training data and adopt MSE loss and DICE loss \cite{milletari2016v} in the second stage. Then we train the BSA module with synthetic images and segmentation masks with pixel-wise Binary Cross Entropy loss independently for 100 epochs. Finally, we jointly train the whole network with the same alternating strategy as mentioned before for 100 epochs. 
For BladeNet training, We use a crop size of $192\times192$ for REDS and REDS+VOC synthetic images, $256\times256$ for LODE images. We use data augmentation with random 90 degree rotation and flip. For evaluation, we use whole images without any resizing and cropping.

\subsection{Detailed Analysis for LBP and BSA Modules}

\begin{table}

\caption{Detailed Analysis for LBP and BSA modules.}

\begin{center}
\begin{tabular}{p{1.1cm}|p{0.7cm}|p{0.8cm}|p{0.7cm}|p{0.8cm}|p{0.7cm}|p{0.8cm}}
\hline
\multirow{2}{*}{Methods} & \multicolumn{2}{c|}{Full} & \multicolumn{2}{c|}{$Threshold $} & \multicolumn{2}{c}{$Threshold $} \\ 
& \multicolumn{2}{c|}{Image} & \multicolumn{2}{c|}{$\leq 0.25$} & \multicolumn{2}{c}{ $>0.25$} \\ \cline{2-7} 
 & PSNR & SSIM & PSNR & SSIM & PSNR & SSIM\\
\hline
\hline
None  & 33.63  & 0.9489 & 61.73 & 0.9981 & 34.05  & 0.9510 \\
LBP & 33.87  & 0.9460 & 62.17  & 0.9979 & 34.29  & 0.9484 \\
BSA & 33.90  & 0.9482 & 60.97  & 0.9979 & 34.40  & 0.9505  \\
L+B & 34.15  & 0.9494 & 62.08  & 0.9981 & 34.59  & 0.9515 \\ \hline

\end{tabular}
\end{center}
\vspace{-10pt}
\label{tab:module_result_sup}
\end{table}

To dig deeper into the local deblurring problem, we investigate the LBP and BSA's performance on regions with different motion. To be specific, we evaluate the performance on small-motion regions (optical flow norm $\leq 0.25$, likely to contain static background) and large motion regions (optical flow norm $> 0.25$, likely to contain moving objects) respectively. From Table \ref{tab:module_result_sup} we observe that both LBP and BSA yields large improvements on the large motion regions, indicating they are particularly beneficial for local motion deblurring. Besides, while BSA works better on large-motion regions and yields superior overall performance, LBP is able to protect those small-motion regions from being over-deblurred. 

\subsection{More Visualization Results}
\paragraph{Visual comparison of with and without LBS module.} We provide more qualitative results to verify the effectiveness of LBS. As illustrated in Fig \ref{fig:LBP_compare_sup}, we observe that LBS can bring significant visual improvement for all baselines, resulting in sharper and more natural results. For instance, the bicycle handle in the first row, the head in the second row, the letters in the third row, the chinese characters in the last row all become clearer and contain less artifact compared to the baseline trained only on REDS. While most global deblurring methods perform poorly on the local deblurring problem, we analyze that the main reason may be that most previous global deblurring datasets are insufficient for solving all kinds of motion deblurring. On the other hand, the proposed LBS is universally beneficial for all deblurring methods and improves them significantly.

\paragraph{Visual comparison of with and without LBP/BSA module.}
We compare the qualitative results of BladeNet without LBP/BSA and the standard BladeNet. The result indicates that both LBP and BSA are indispensable for the local deblurring. The result in Fig \ref{fig:LBP_BSA_sup} shows that the while the LBS generated synthetic data makes the original blurred images clearer, the LBP and BSA are able to further compensate the local deblurring effect. For example, the leg of the leftmost running person on the upper left, the bicycle wheels on the upper right, the wheels of the car on the bottom left, the basketball on the bottom right, all become clearer when equipped with the LBP and BSA modules. The result shows that with the synthetic data and segmentation information, the particularly designed network structures also contribute greatly to the local deblurring.

\paragraph{Visual comparison of LODE finetuning.}
We compare the results BladeNet and BladeNet finetuned on LODE. The result in Fig \ref{fig:finetuned_result_sup} demonstrates that while BladeNet yields impressive local deblurring results, finetuning on LODE is beneficial for further image quality improvement.

\end{document}